\documentclass[letterpaper, 10 pt, conference]{ieeeconf}  

\IEEEoverridecommandlockouts                              

\usepackage{amsmath}
\usepackage{comment}  
\usepackage{diagbox} 

\usepackage{graphicx} 
\usepackage{float} 
\usepackage{subfigure} 
\usepackage{multirow}
\usepackage{makebox}
\usepackage{booktabs}

\usepackage{hyperref}
\usepackage{cleveref}  
\crefname{figure}{fig.}{figures}
\Crefname{figure}{Fig.}{Figures}
\crefname{table}{tab.}{tables}
\Crefname{table}{Tab.}{Tables}
\crefname{equation}{eq.}{equations}
\Crefname{equation}{Eq.}{Equations}

\usepackage[T1]{fontenc}
\usepackage{colortbl} 
\usepackage[table]{xcolor}
\usepackage{amssymb} 
\usepackage{array}

\usepackage{pifont}  
\usepackage{bbding}  

\usepackage[hang,flushmargin]{footmisc} 
\usepackage{tikz} 
\usepackage{threeparttable}

\def\onedot{.}
\def\eg{\emph{e.g}\onedot} 
\def\ie{\emph{i.e}\onedot} 
 
 \def\vs{\emph{$v.s$}\onedot\ }

\def\etal{\emph{et al}\onedot\ }
\makeatother
\title{\LARGE \bf
BEV$^2$PR: BEV-Enhanced Visual Place Recognition with Structural Cues
}

\author{Fudong Ge$^{1,2\dag}$, Yiwei Zhang$^{1,2\dag}$, Shuhan Shen$^{1,2}$, Weiming Hu$^{1,2,3}$, Yue Wang$^{4}$, Jin Gao$^{1,2*}$ 
\thanks{$^{\dag}$Equal contribution. $^{*}$Corresponding author (jin.gao@nlpr.ia.ac.cn).}
\thanks{$^{1}$State Key Laboratory of Multimodal Artificial Intelligence Systems (MAIS), CASIA. $^{2}$School of Artificial Intelligence, University of Chinese Academy of Sciences. $^{3}$School of Information Science and Technology, ShanghaiTech University. $^{4}$State Key Laboratory of Industrial Control Technology and Institute of Cyber-Systems and Control, Zhejiang University.}
}

\begin{document}

\maketitle
\thispagestyle{empty}
\pagestyle{empty}

\begin{abstract}

In this paper, we propose a new image-based visual place recognition (VPR) framework by exploiting the structural cues in bird’s-eye view (BEV) from a single monocular camera. The motivation arises from two key observations about place recognition methods based on both appearance and structure: 1) For the methods relying on LiDAR sensors, the integration of LiDAR in robotic systems has led to increased expenses, while the alignment of data between different sensors is also a major challenge. 2) Other image-/camera-based methods, involving integrating RGB images and their derived variants (\eg, pseudo depth images, pseudo 3D point clouds), exhibit several limitations, such as the failure to effectively exploit the explicit spatial relationships between different objects. To tackle the above issues, we design a new BEV-enhanced VPR framework, namely BEV$^2$PR, generating a composite descriptor with both visual cues and spatial awareness based on a single camera. The key points lie in: 1) We use BEV features as an explicit source of structural knowledge in constructing global features. 2) The lower layers of the pre-trained backbone from BEV generation are shared for visual and structural streams in VPR, facilitating the learning of fine-grained local features in the visual stream. 3) The complementary visual and structural features can jointly enhance VPR performance. Our BEV$^2$PR framework enables consistent performance improvements over several popular aggregation modules for RGB global features. The experiments on our collected VPR-NuScenes dataset demonstrate an absolute gain of 2.47\% on Recall@1 for the strong Conv-AP baseline to achieve the best performance in our setting, and notably, a 18.06\% gain on the hard set. The code and dataset will be available at \textit{https://github.com/FudongGe/BEV2PR}.

\end{abstract}

\section{INTRODUCTION}

Visual Place Recognition (VPR) plays a crucial role in robotics and autonomous driving, typically framed as an image retrieval task \cite{lowry2015visual, masone2021survey, zhang2021visual}. The primary objective of a VPR system is to ascertain the spatial position of a provided query image, involving the initial extraction of its visual information into a compact representation, and a subsequent comparison with a reference database with known geolocations. These image-based methods are susceptible to challenges such as varying illumination and weather, attributed to the intrinsic characteristics of camera sensors \cite{luo2023bevplace}. Therefore, a question arises naturally: \textit{How to learn a robust representation for VPR based solely on cameras?}

\begin{figure}[] 
\centering 
\includegraphics[width=0.46\textwidth]{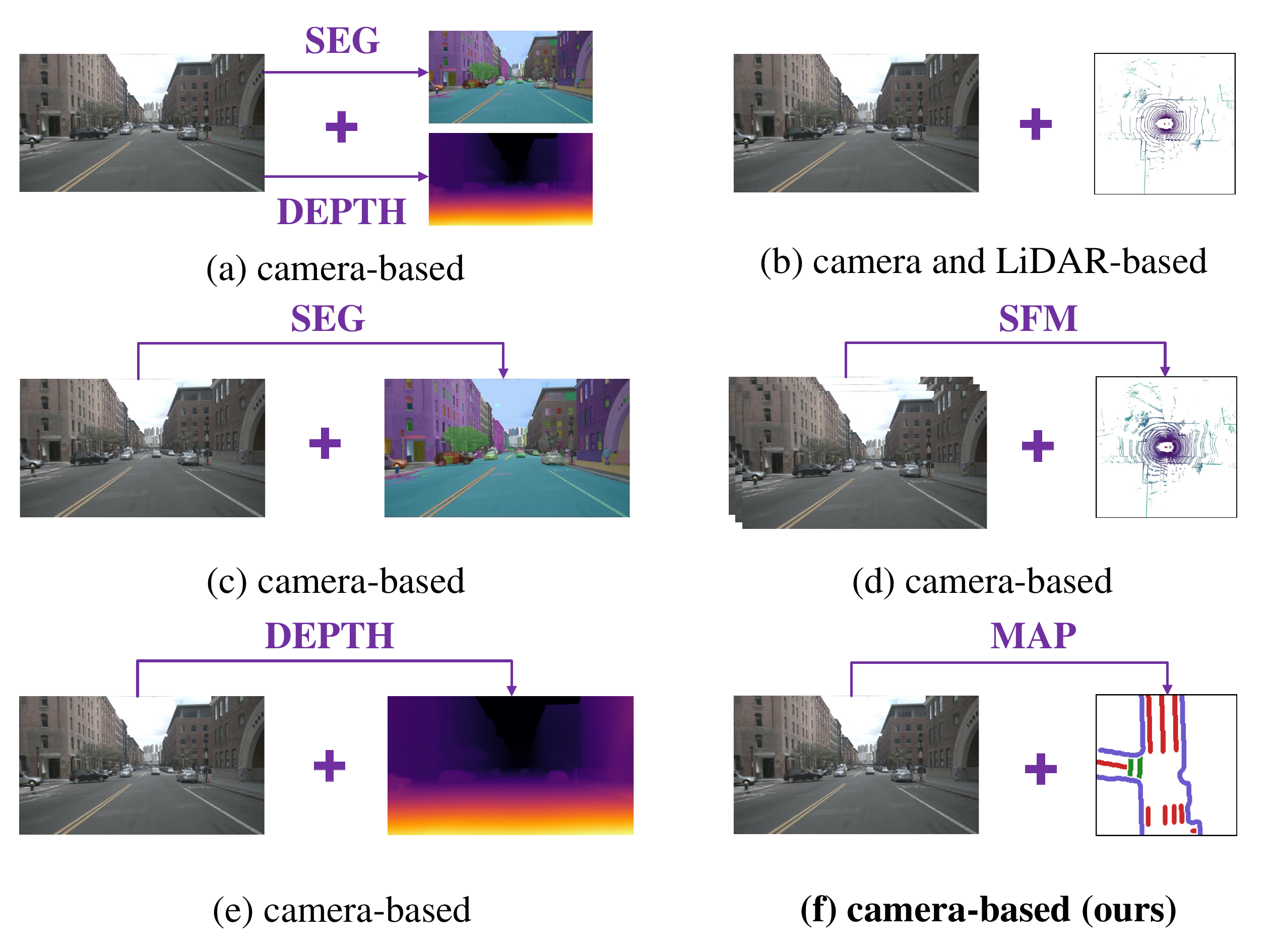} 
\setlength{\abovecaptionskip}{-6pt} 
\caption{Schematic diagram of methods based on both appearance and structure using camera and/or LiDAR sensors: a) methods using RGB, segmentation and depth images, b) methods using RGB images from camera and 3D point clouds from LiDAR, c) methods using RGB and segmentation images, d) methods using RGB images and 3D point clouds reconstructed from multiple RGB images, e) methods using RGB and depth images, f) (ours) using RGB images and BEV generated from RGB images.} 
\label{Fig.1} 
\vspace{-0.8cm}
\end{figure}

After reviewing the progress in place recognition methods, we find that the research based on both appearance and structure is a hotter field \cite{komorowski2021minkloc++, lai2022adafusion, zhou2023lcpr}. Despite the primary focus of this paper being on VPR tasks, we also introduce some methods related to LiDAR to emphasize the importance of structural features. As depicted in \Cref{Fig.1}, existing methods based on both appearance and structure can be divided into approaches using both camera and LiDAR sensors \cite{komorowski2021minkloc++, lai2022adafusion, zhou2023lcpr}, \ie, (b), and camera-based approaches \cite{hu2020dasgil, oertel2020augmenting, shen2023structvpr}, \ie, (a), (c), (d) and (e). Furthermore, the structural information sources of camera-based methods include 2D segmentation images, reconstructed 3D point clouds, 2D depth images, thus (c) \vs (d) \vs (e) is also a focus worth analyzing.

Upon examining the current methods, it is noted that each possesses its own advantages and disadvantages. The methods based on both camera and LiDAR in \Cref{Fig.1}(b), which utilize the complementarity of multimodal raw information, face issues involving expensive equipments, as well as the calibration and alignment of different modal data \cite{komorowski2021minkloc++, lai2022adafusion, zhou2023lcpr}. Vehicles not equipped with LiDAR cannot deploy such algorithms. The SfM (Structure-from-Motion)-based methods with pseudo 3D point clouds reconstructed from RGB images, \ie, \Cref{Fig.1}(d), avoids the expense of using LiDAR. However, this reconstruction process is relatively complex, and the data required for inference are more challenging to handle than conventional pixel images \cite{oertel2020augmenting}. Beneficial from current segmentation models \cite{kirillov2023segment}, SEG-based methods in \Cref{Fig.1}(c) use segmentation images to enhance structural knowledge of global features. However, segmentation images are still in the x-y 2D plane without the implicit structure - depth, resulting in the loss of key information. \Cref{Fig.1}(e) provides depth information, while depth images contain harmful noise from dynamic objects, and lack explicit spatial relationships between different objects. The limitations of the current methods raise a question: \textit{Could we integrate explicit depth and spatial relationships as well as RGB information into global features using images as input during inference?}

\begin{figure*}[t] 
\centering 
\includegraphics[width=1\textwidth]{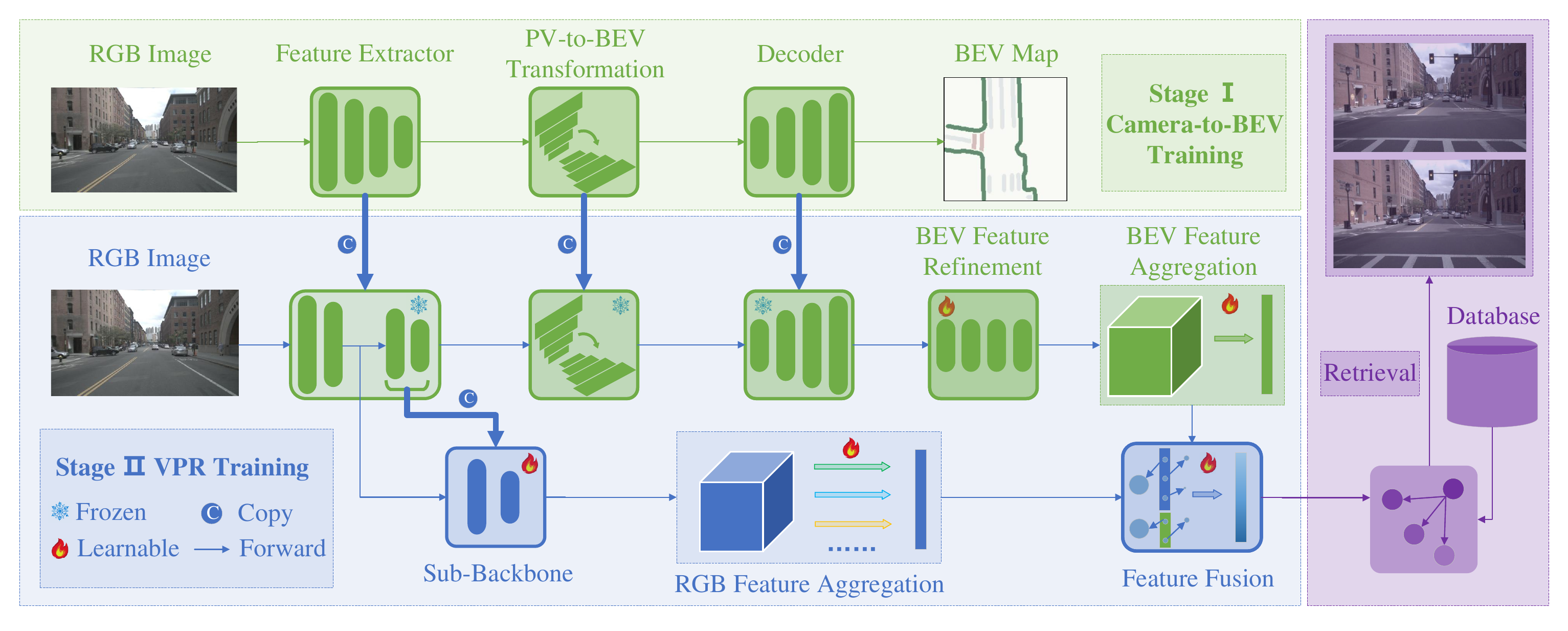} 
\setlength{\abovecaptionskip}{-20pt} 
\caption{Overview of our proposed pipeline. (1) In Stage \uppercase\expandafter{\romannumeral1}, we pre-train a BEV generation model using a front-view image to extract BEV features as a more explicit source of structural knowledge in the global feature. (2) In Stage \uppercase\expandafter{\romannumeral2}, we first copy the modules of the BEV model as part of the structural stream and freeze them to introduce BEV features into VPR, then crop its backbone into two parts, with the former serving as the bottom backbone shared with the visual stream and the latter copied as the sub-backbone of visual stream and then unfrozen. After that, any aggregation module for RGB global features is inserted into the visual stream and a relatively simple GeM module is used in the structural stream, followed by a feature fusion operation. (3) Finally, the nearest neighbor search is used to retrieval the top-$k$ images.} 
\label{Fig.2} 
\vspace{-0.7cm}
\end{figure*}

To tackle this issue, we propose our solution - \textbf{RGB and BEV fusion}, inspired by the widespread application of bird's-eye view (BEV) representations in 3D perception tasks, where BEV exhibits outstanding performance in clearly depicting the relative positions. The challenge lies in \textit{how to integrate BEV features into global features and leverage knowledge from the structural stream to enhance the visual stream.} Therefore, we propose \textbf{BEV$^2$PR}, a new architecture simultaneously constructing semantic map to model spatial relationships in the \textbf{BEV} frame, and generating a composite descriptor with both visual cues and spatial awareness for \textbf{VPR}. Specifically, we first pre-train a BEV generation model to extract BEV features as a more explicit source of structural knowledge. And we copy the modules of this model as part of the structural stream and freeze them to introduce BEV features into VPR, then crop its backbone into two parts, with the former serving as the bottom backbone shared with visual stream and the latter copied as the sub-backbone of visual stream and then unfrozen. The details are shown in \Cref{Fig.2}. After that, any aggregation module for RGB global features is inserted into the visual stream, followed by a fusion operation. In particular, our framework only uses an image as input, without involving other sensors or complex training processes, making it generalized for a wide range of autonomous vehicles.

Our main contributions can be highlighted as follows: 1)  \textbf{Data Module}: We introduce VPR-NuScenes to distinguish between simple and challenging scenes to understand the respective strengths of visual and structural representations. 2) \textbf{Architecture}: We propose BEV$^2$PR to simultaneously construct semantic map to model spatial relationships in the BEV frame, and generate a composite descriptor with both visual cues and spatial awareness for VPR. 3) \textbf{Experimental Results}: We conduct extensive experiments to evaluate the improvement effect of our framework on different appearance-based methods, particularly in scenes characterized by significant appearance variations.

\section{RELATED WORK}
The related work involves appearance-based place recognition as well as place recognition based on both appearance and structure. The former focuses on studying the semantic and texture information of RGB images, while the latter introduces structural information on the basis of the former.

\vspace{-0.2cm}
\subsection{Place Recognition with Appearance}
The appearance-based place recognition concentrates on constructing better image representations including global or local descriptors for retrieval, where global descriptors can be generated through direct extraction or by aggregating local descriptors \cite{shen2023structvpr}. As a representative of aggregation algorithm, NetVLAD \cite{arandjelovic2016netvlad} is a trainable variant of VLAD, softly allocating local features to a set of learned clusters. Based on it, many variants have been inspired, such as SPE-NetVLAD \cite{yu2019spatial} and Gated NetVLAD \cite{zhang2021vector}.

Several methods emphasize identifying key regions within feature maps. One notable technique, GeM \cite{radenovic2018fine}, represents a learnable variant of global pooling, expanding upon which, Berton \etal introduce CosPlace \cite{berton2022rethinking} that combines GeM with a linear layer, exhibiting strong performance. TransVPR \cite{wang2022transvpr} by Wang \etal integrates CNN with Transformer by multi-head self-attention mechanism to infuse attention within output tokens from Transformer encoder. In \cite{ali2022gsv}, Ali-bey \etal focus on processing the high-level features and propose Conv-AP, which implements channel-wise pooling on the features followed by spatial-wise adaptive pooling, achieving state-of-the-art results on multiple benchmarks.

\vspace{-0.1cm}
\subsection{Place Recognition with both Appearance and Structure}

\textbf{Place Recognition Relying on LiDAR.} Considering the robustness of structural features in some environments, scholars have attempted to introduce fusion technologies of point cloud and image into place recognition. PIC-Net \cite{lu2020pic} by Lu \etal employs global channel attention to enhance the interaction between point cloud and image features, along with a spatial attention-based VLAD to select the discriminative points and pixels. Zhou \etal develop LCPR \cite{zhou2023lcpr}, a multi-scale network leveraging self-attention mechanism to correlate panoramic features across different modalities. OneShot \cite{ratz2020oneshot} by Ratz \etal projects the segments from point clouds onto images to facilitate feature extraction through 2D and 3D CNN. CORAL \cite{pan2021coral} by Pan \etal first builds an elevation image from LiDAR scans, and then augments it with projected visual features to generate a fusion descriptor.

\textbf{Place Recognition Independent of LiDAR.} Another path relies solely on cameras. Oertel \etal \cite{oertel2020augmenting} use structural features extracted from image sequences through vision-based SfM to augment VPR. Shen \etal \cite{shen2023structvpr} leverage segmentation images obtained from a pre-trained segmentation model to augment the structural comprehension embedded within global representations, achieving state-of-the-art performance. Based on knowledge transfer and adversarial learning, Qin \etal \cite{qin2021structure} propose a structure-aware feature disentanglement network, called SFDNet, which leverages probabilistic knowledge transfer to transmit features from Canny edge detector to structure encoder. DASGIL \cite{hu2020dasgil} by Hu \etal uses a multi-task architecture to integrate geometric and semantic knowledge into multi-scale global representations.

In this paper, we construct semantic map to model spatial relationships within the BEV frame, and for the first time, adopt a late-fusion approach of RGB and BEV features to enhance VPR, benefitting from the explicit spatial relationship modeling  between different objects in BEV generation.

\section{METHODOLOGY}

Considering the rich and explicit structural knowledge inherent in BEV representations, we leverage BEV features to infuse structural information into global retrieval. The core concept behind BEV$^2$PR focuses on the static features in BEV images and the explicit spatial relationships between different objects, as well as the single-input but dual-modal processing method. BEV$^2$PR involves two training stages: Camera-to-BEV training and VPR training shown in \Cref{Fig.2}.

\vspace{-0.2cm}
\subsection{Overview}

Given an input RGB image, a BEV generation model is pre-trained in the first training stage, by which BEV features can be obtained to assist in VPR. In the second training stage, the BEV model is frozen and its backbone is cropped into two sequential parts. Finally, we insert any global feature aggregation module into the visual stream, followed by a fusion operation to obtain a composite descriptor.

\vspace{-0.2cm}
\subsection{BEV Map Generation}

\textbf{Semantic Class Selection.} In various BEV images, there are different numbers of semantic instances, making the selection of appropriate types and quantities significant. A BEV map is represented by a $C \times H_s \times W_s$ tensor, where $C$ represents the number of semantic classes, and $H_s \times W_s$ corresponds to the area in front of the ego vehicle. Within this tensor, regions belonging to the $c^{th}$ class are marked with positive values in the corresponding $c^{th}$ channel, while those in other channels are set to zero. And, in the context of VPR, we improve model convergence and accuracy by only using static classes as the target to train the BEV generation model. This technique draws inspiration from the fact that each semantic class plays a different role in VPR, and further human brain’s attention mechanism, which concentrates on iconic static elements or important regions, such as roads and buildings, while ignoring dynamic ones similar to noise.

\textbf{Camera-to-BEV Model.} In this section, we establish a BEV generation model with only a front-view image as input. First, an image is embedded by a 2D CNN to get its feature map. And then, we transform the feature map from 2D to 3D using a similar method with \cite{philion2020lift}, that is 1) predict grid-wise depth distribution with equal spacing on 2D features, 2) ‘lift’ the 2D features to voxel space based on depth, obtaining pseudo point cloud features. Finally, a pooling operation similar to LiDAR-based methods is implemented to flatten 3D point cloud features into 2D BEV features.

\vspace{-0.2cm}
\subsection{Structural Knowledge Extraction}

\textbf{Shared Bottom Backbone.} Due to the insensitivity of global features to small targets in the raw visual stream, we adopt a shared bottom backbone instead of the architecture of separately processing input images in visual and structural streams. To specify, we crop the backbone of BEV generation model into two sequential parts, with the former serving as a bottom backbone shared with the visual stream, so as to prompt visual stream to learn fine-grained local features generated by the BEV semantic segmentation, meanwhile, the latter is copied as the specific sub-backbone of visual stream and unfrozen to capture important features for VPR.

\textbf{Feature Refinement Module.} Considering that the BEV feature map operates at a higher semantic level compared to the RGB image and needs to be refined to denoise and adapt to VPR, we adopt a part of depth stream in MobileSal \cite{wu2021mobilesal} as the feature refinement module in the structural stream. Specifically, there are totally five stages with the same strides, the last of which is modified to refine BEV features.

\vspace{-0.2cm}
\subsection{Preliminary Feature Aggregation}
In our work, we evaluate multiple methods to verify the improvement effect of BEV features on them. The feature in the visual stream is denoted as $\mathcal{X} \subseteq \mathcal{R}^{K \times H \times W}$, where $K$ represents the number of features maps. And $\mathcal{X}_k$ is defined as the 2D feature map of $H \times W$ dimensions, where $k \in \{1, \dots, K\}$. Let $f_v$ be the output representation.

\textbullet\  SPoC \cite{7410507}: This technique involves aggregating the features extracted from CNN through sum pooling, \ie,
\begin{equation}
\label{Eq.1}
f_v = [f_v^{(1)}, f_v^{(2)}, \dots, f_v^{(K)}]^T,
\end{equation}
\begin{equation}
f_v^{(k)} = \frac{1}{|\mathcal{X}_k|}\sum_{x \in \mathcal{X}_k}x.
\end{equation}

\textbullet\  NetVLAD \cite{arandjelovic2016netvlad}: As a representative of aggregation algorithm, it is a trainable variant of VLAD, softly allocating
local features to a set of learned clusters, defined as
\begin{equation}
f_v = [f_v^{(1)}, f_v^{(2)}, \dots, f_v^{(R)}]^T,
\end{equation}
where $R$ represents the number of cluster centers, and
\begin{equation}
f_v^{(r)} = \sum_{i=1}^{N}\frac{e^{w_r^Tx_i+b_r}}{\sum_{r'}e^{w_{r'}^Tx_i+b_{r'}}}(x_i-c_r),
\end{equation}
where $x_i$ represents the $i^{th}$ descriptor, $c_r$ represents the $r^{th}$ cluster center, $w_r$ and $b_r$ are sets of trainable parameters for each cluster, $N=H \times W$ represents the total number of features. Refer to \cite{arandjelovic2016netvlad} for more details.

\textbullet\  GeM \cite{radenovic2018fine}: This method implements a parametric generalized-mean mechanism, which can assign a shared parameter per feature map. Owing to the differentiable nature of pooling operation, the parameters can be learned as part of the back-propagation. The global feature is defined the same as \Cref{Eq.1}, but the local feature is denoted as
\begin{equation}
f_v^{(k)} = (\frac{1}{|\mathcal{X}_k|}\sum_{x \in \mathcal{X}_k}x^{p_k})^\frac{1}{p_k},
\end{equation}
where $p$ represents the learnable parameter. 

\textbullet\  Conv-AP \cite{ali2022gsv}: This is a fully convolutional feature aggregation technique, which implements channel-wise pooling on feature maps, coupled with spatial-wise adaptive pooling, allowing for significant configurability in output dimensionality. Conv-AP is defined as
\begin{equation}
f_v = AAP_{s_1 \times s_2}(Conv_{1 \times 1} (\mathcal{X})),
\end{equation}
where $AAP$ represents the adaptive average pooling, $s_1 \times s_2$ represents the number of spatial sub-regions, $Conv_{1 \times 1}$ represents the $1 \times 1$ convolution.

\textbullet\  EigenPlaces \cite{berton2023eigenplaces}: As a variant that builds on GeM aggregator, EigenPlaces utilizes a GeM pooling and a fully connected (FC) layer with output dimension 512, defined as
\begin{equation}
f_v = FC(GeM(\mathcal{X})).
\end{equation}

\begin{figure*}[t] 
\centering 
\includegraphics[width=1\textwidth]{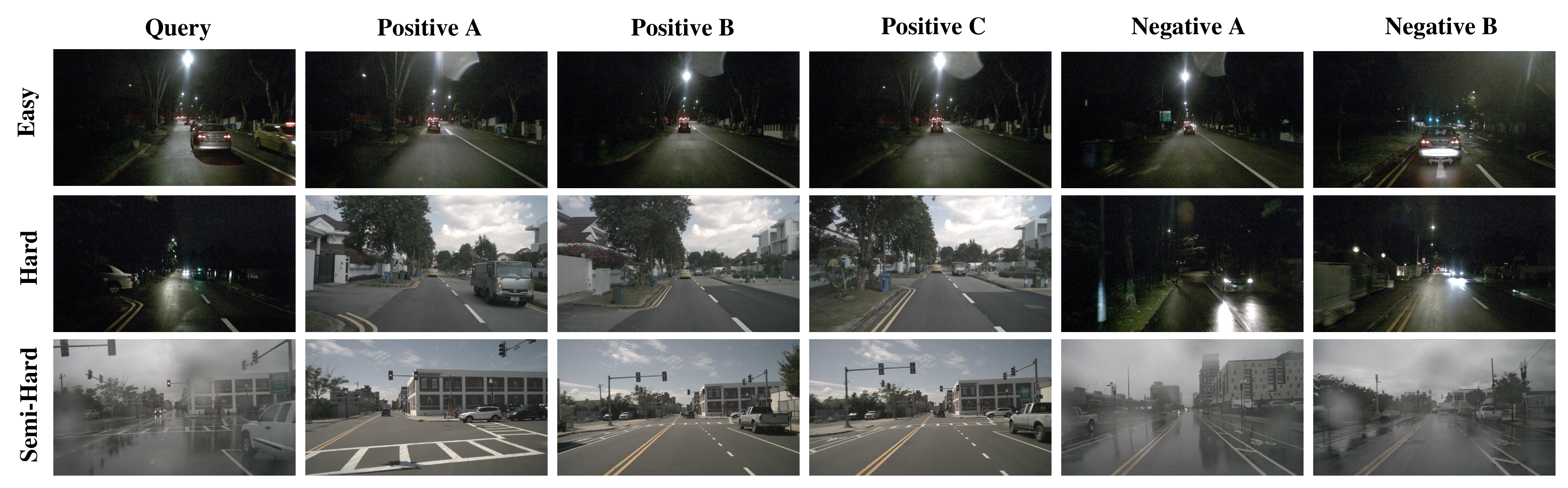} 
\setlength{\abovecaptionskip}{-14pt} 
\caption{Example of samples with different recall difficulties on our collected VPR-NuScenes dataset.} 
\label{Fig.3} 
\vspace{-0.3cm}
\end{figure*}

\textbullet\  MixVPR \cite{ali2023mixvpr}: This work focuses on an isotropic all-MLP architecture, and utilizes feature maps extracted from a pre-trained backbone as an aggregation of global descriptors to integrate the global interaction among elements within each feature map through a cascade of feature mixing.
First, $\mathcal{X}$ is considered as a set of 2D $H \times W$ features, \ie, $\mathcal{X} = \{\mathcal{X}_1, \mathcal{X}_2, \dots, \mathcal{X}_K\}$, followed by a flattening operation, resulting in feature maps $\mathcal{X} \subseteq \mathcal{R}^{N \times K}$, where  $N = H \times W$. Then, it is fed into $I$ Feature Mixer blocks, \ie,
\begin{equation}
\mathcal{Z} = FM_I(FM_{I-1}(\dots FM_1(\mathcal{X}))).
\end{equation}
Finally, two fully connected layers are used to reduce its dimension depth-wise (channel-wise) then row-wise, \ie,
\begin{equation}
\mathcal{Z}' = \mathcal{W}_{d}(Transpose(\mathcal{Z})),
\end{equation}
\begin{equation}
f_v = \mathcal{W}_{r}(Transpose(\mathcal{Z}')),
\end{equation}
where $\mathcal{W}_d$ and $\mathcal{W}_r$ are the weights of two layers.

Finally, $f_v$ is flattened and $L_2$-normalized as usually done in VPR \cite{arandjelovic2016netvlad}, still recorded as $f_v$ for convenience.

\subsection{Feature Fusion}


Here, we use a weighted concatenation with two learnable scalar weights $w_v$ and $w_s$ to rescale each descriptor in two streams, and the feature aggregation module in the structural stream is set to use GeM with the output representation denoted as $f_s$. Then the final descriptor is given by
\begin{equation}
f = concat(w_vf_v, w_sf_s).
\end{equation}

\subsection{Network Training}

We train our compound network in a two-stage manner. The loss function in the first stage fully refers to existing work \cite{philion2020lift, li2022hdmapnet}. For the second stage, each training iteration involves a mini-batch $(q, P_q, N_q)$ consisting of one query sample $q$ as well as multiple positive matches $P_q$ and negative matches $P_n$ for it. Following \cite{arandjelovic2016netvlad}, we adopt the triplet margin loss for each tuple, \ie,
\begin{equation}
\mathcal L_{ij} = max(d(l(q), l(p_i)) - d(l(q), l(n_j)) + m, 0),
\end{equation}
where $p_i$ and $n_j$ represent the $i^{th}$ positive sample and the $j^{th}$ negative sample for $q$ in a mini-batch, margin $m$ is a constant, $d(\cdot)$ computes the Euclidean distance of two descriptors, $l(\cdot)$ is defined as the function mapping input to its descriptor. Then, the loss function of each descriptor is defined as
\begin{equation}
\mathcal L_t = \frac{1}{N_{\text{pos}}N_{\text{neg}}} \sum_{i=1}^{N_{\text{pos}}} \sum_{j=1}^{N_{\text{neg}}} \mathcal L_{ij},
\end{equation}
where $N_{pos}$ and $N_{neg}$ represent the number of positive and negative samples in each mini-batch, $t\!\in\{F,V,S\}$ represent the type of three descriptors, \ie, fusion descriptor loss $\mathcal L_F$, visual descriptor loss $\mathcal L_V$, structural descriptor loss $\mathcal L_S$. Considering that the network tends to overfit in one single domain during training \cite{komorowski2021minkloc++}, we use a multi-head loss function for final back-propagation, denoted as
\begin{equation}
\mathcal L_{vpr} = \alpha\mathcal L_F + \beta\mathcal L_V + \gamma\mathcal L_S,
\end{equation}
where $\alpha$, $\beta$, $\gamma$ are constants determined by experiments.

\section{EXPERIMENTS}

\vspace{-4pt}
\subsection{Dataset}

Although there are multiple datasets available for VPR in challenging conditions, a deficiency is observed in annotated information for BEV semantic segmentation maps among them. Given our method's dependence on BEV features for extracting structural information, evaluations are performed on the NuScenes \cite{caesar2020nuscenes}, a public large-scale dataset for autonomous driving equipped with an entire suite of multi-modal sensors, including 6 cameras, 1 LiDAR, 5 radars, 1 IMU and 1 GPS. Among all the 1k scenes, each comprising a 20s-long sequence of consecutive frames, the trainval split includes 850 annotated scenes suitable for BEV generation, while the test split with 150 scenes has no annotations. Thus, only the annotated 850 scenes are eligible for our method.

\textbf{Partitioning of Positive and Negative Samples.} Given the spatial proximity, the limited variations in camera angles and the minimal changes in appearance between consecutive frames, treating consecutive frames as positive samples proves excessively facile and lacks practical significance. Consequently, images within the same scene are not regarded as positive samples to one another. The selection of negative samples is straightforward: for each query sample, samples other than its positive samples and consecutive frames are defined as negative samples.

\begin{table}[t]
\centering
\setlength{\abovecaptionskip}{-1pt} 
\caption{Definition of Sample Recall Difficulty.}
\label{Tab.1} 
\setlength{\tabcolsep}{6pt} 
\renewcommand{\arraystretch}{1.2}
\scalebox{0.89}{
\begin{tabular}{c|cccc}
\hline
\diagbox[dir=NW, width=2cm, height=18pt]{Query}{Positive} & Day & Night & Day\&Rain & Night\&Rain \\ \hline
Day         & Easy      & Hard      & Semi-Hard & Hard        \\
Night       & Hard      & Easy      & Hard      & Semi-Hard   \\
Day\&Rain   & Semi-Hard & Hard      & Easy      & Hard        \\
Night\&Rain & Hard      & Semi-Hard & Hard      & Easy        \\
\hline
\end{tabular}
}
\vspace{-14pt}
\end{table}

The exclusive reliance on GPS for distinguishing between positive and negative samples presents challenges \cite{ge2020self}. Images captured in close geographical location may fail to depict identical scenes when oriented differently, leading to pseudo positive samples. And variations in camera position while photographing the same site may introduce the possibility of pseudo negative samples. Due to the limitations in reflecting place position with camera coordinates, we employ image coordinates rather than camera coordinates to determine positive samples corresponding to a query sample.

In this work, the position of an image $p_{img}$ is defined as $25m$ in front of the camera, aligned with the center of BEV image. After comparison, the visual effect of the positive images determined by this position is the best. In order to obtain more reasonable initial positive and negative samples, each image is represented by a binary tuple $<$$p_{img}, v$$>$, where, $v$ represents the directional vector of camera-captured images, denoted as
\begin{equation}
v = p_{img} - p_{cam},
\end{equation}
where $p_{cam}$ represents the position of the camera.

Then, we apply the Euclidean distance metric to calculate the distance between two images and those with a distance less than the threshold from other images are regarded as candidate positive samples, followed by angle calculation between the direction vectors of the query and positive samples to determine the final positive samples, \ie,
\begin{equation}
\theta = \arccos(\frac{v_q \cdot v_{cp}}{\|v_q\|\|v_{cp}\|}),
\end{equation}
where $v_q$ and $v_{cp}$ represent the direction vectors of the query image and the candidate positive images respectively. To specify, the distance and angular thresholds for positive samples are set to $10m$ and $30^\circ$ respectively.

Furthermore, in order to more clearly analyze the pros and cons of different modalities, we classify samples based on the appearance differences between query and positive images, and define it as follows (refer to \Cref{Tab.1} for details):

\begin{table}[t]
\centering
\setlength{\abovecaptionskip}{-1pt} 
\caption{Statistics of our data organization.}
\setlength{\tabcolsep}{4pt} 
\renewcommand{\arraystretch}{1.2}
\scalebox{0.89}{
    \begin{threeparttable} 
        \begin{tabular}{>{\centering\arraybackslash}p{0.4cm}|
        >{\centering\arraybackslash}p{0.8cm}>{\centering\arraybackslash}p{0.6cm}>{\centering\arraybackslash}p{0.6cm}>{\centering\arraybackslash}p{0.6cm}<{\centering}|
        >{\centering\arraybackslash}p{0.7cm}>{\centering\arraybackslash}p{0.6cm}>{\centering\arraybackslash}p{0.4cm}|
        >{\centering\arraybackslash}p{0.6cm}>{\centering\arraybackslash}p{0.8cm}}
        \hline
        \multirow{2}{*}{} & \multirow{2}{*}{D} & \multirow{2}{*}{N} & \multirow{2}{*}{D\&R} & \multirow{2}{*}{N\&R} & \multirow{2}{*}{E} & \multirow{2}{*}{SH} & \multirow{2}{*}{H} & \multicolumn{2}{c}{Total} \\ \cline{9-10} 
        &  &  &  &  &  &  &  & Scene & Sample \\ \hline
        N$_{\text{tr}}$ & 14799 & 1239 & 3887 & 261 & 17335 & 1997 & 854 & 583 & 20186 \\
        N$_{\text{te}}$ & 5426 & 539 & 1176 & 124 & 5960 & 956 &349 &235 & 7265 \\ \hline
        \end{tabular}
        \begin{tablenotes} 
            \item N$_{\text{tr}}$ - N$_{\text{train}}$, N$_{\text{te}}$ - N$_{\text{test}};$
        \end{tablenotes} 
        \begin{tablenotes} 
            \item D - Day, N - Night, D\&R - Day\&Rain, N\&R - Night\&Rain; 
        \end{tablenotes} 
        \begin{tablenotes} 
            \item E - Easy, SH - Semi-Hard, H - Hard.
        \end{tablenotes} 
    \end{threeparttable} 
}
\label{Tab.2} 
\vspace{-16pt}
\end{table}

\textbullet\ Hard Sample: significant illumination differences. For instance, the query image is taken during daylight, while \textbf{all} positive images are captured at nighttime.

\textbullet\ Semi-Hard Sample: substantial contrast in image clarity due to precipitation. For example, the query is taken during the daytime, but \textbf{all} positive images are acquired in rain.

\textbullet\ Easy Sample: minimal appearance differences.

\textbf{Partition of Training and Test Sets.} Considering that NuScenes only involves four regions, each with data collected in relatively concentrated periods, we opt for a holistic partitioning approach for data across all areas. In VPR, the research on recalling from day to night, as well as from night to rain presents more challenges, but holds greater significance. Therefore, we aim to achieve a uniform distribution of such data in both the training and testing sets. This data distribution design facilitates a more comprehensive evaluation of the model's adaptability to complex environmental changes, enhancing the model's generalization performance.

We first establish an undirected graph with scenes as nodes and similarities between scenes as edge weights, where the scene similarity is defined as the proportion of each query image in one scene that has a corresponding positive image appearing in another scene. Specifically, if all query images have positive images in the same scene, the scene similarity is set to 1. After processing all the scenes, there are 32 isolated scenes without connecting to any other scenes, making them unusable for VPR but usable for BEV generation. Finally, we leverage the official API of NuScenes dataset to acquire scene labels, such as "night" or "day", and based on which, the data is manually balanced, generating the partitioning results in \Cref{Tab.2}. \textit{Note that there is no any instance and scene overlap between the training set and the test set}. This dataset is called \textbf{VPR-NuScenes}.

\vspace{-0.2cm}
\subsection{Implementation Details}

\textbf{Architecture.} We implement BEV$^2$PR in PyTorch framework and reproduce those methods with open-source codes for a fair comparison, including SPoC \cite{7410507}, NetVLAD \cite{arandjelovic2016netvlad}, GeM \cite{radenovic2018fine}, Conv-AP \cite{ali2022gsv}, EigenPlaces \cite{berton2023eigenplaces}, MixVPR \cite{ali2023mixvpr}. For all techniques involving BEV generation and VPR, we use EfficientNet-B0 as the backbone cropped at the last convolutional layer. And the sub-backbone in the visual stream is cropped from the $12^{th}$ block of EfficientNet-B0. Considering the information advantages of different scales, for BEV generation, we adopt a method of upsampling the output of the $16^{th}$ block and concatenating it with the output of the $11^{th}$ block as final extracted features. The Camera-to-BEV model outputs a $50m \times 50m$ map at a $25cm$ resolution with three classes, including lane boundary (Bound.), lane divider (Div.), and pedestrian crossing (P.C.), \ie, $C = 3$. Each input front-view image is resized from $1600 \times 900$ to $704 \times 256$ in all experiments for all models. 

\textbf{Training.} During Stage \uppercase\expandafter{\romannumeral1}, we train the Camera-to-BEV network with EfficientNet-B0 pre-trained on ImageNet for 40 epochs. The Adam optimizer has a weight decay of \(1e^{-7}\) with a learning rate of \(1e^{-3}\). During Stage \uppercase\expandafter{\romannumeral2}, the final descriptor dimension of all methods is set to $640$. We train the whole network using Adam optimizer having a weight decay of \(1e^{-3}\) with a learning rate of \(1e^{-5}\) for $40$ epochs. For each mini-batch, the number of positive samples $N_{pos}=1$, negative samples $N_{neg}$ = 6 and $m = 0.5$ are set for the triplet loss. In addition, the models for the whole dataset and all other subsets use the same hyper-parameters, with $\alpha = 1$, $\beta = 1$, $\gamma = 1$ specifically set in the multi-head loss function.

\begin{table*}[t]
    \vspace{0.2cm}
    \setlength{\abovecaptionskip}{-1pt} 
    \caption{Evaluation of visual place recognition performance.}
    \centering
    \scalebox{0.89}{
    \setlength{\tabcolsep}{5pt} 
    \renewcommand{\arraystretch}{1.20} 
    \begin{threeparttable}
        \begin{tabular}{c|c|c|c|cccccccccccc}
            \hline
        
            Method & Year & Modality & BEV & R@1 & R@5 & R@10 & R$^{\text{E}}$@1 & R$^{\text{E}}$@5 & R$^{\text{E}}$@10 & R$^{\text{H}}$@1 & R$^{\text{H}}$@5 & R$^{\text{H}}$@10 & R$^{\text{SH}}$@1 & R$^{\text{SH}}$@5 & R$^{\text{SH}}$@10 \\ \hline
            
            \multirow{2}{*}{SPoC \cite{7410507}} & \multirow{2}{*}{2015} & \multirow{2}{*}{V} & w/o & 78.44 & 87.45 & 90.13 & 84.76 & 93.07 & 95.39  & 0.00  & 0.00  & 0.00 & 67.47 & 84.00 & 89.96 \\
            
             &  &  & w & 84.04 & 90.44 & 92.27 & 89.70 & 95.14 & 96.39 & 10.60 & 12.89 & 16.62 & 75.52 & 89.44 & 94.14 \\
            
            \cellcolor{lightgray!40}$\Delta$
            & \cellcolor{lightgray!40}-
            & \cellcolor{lightgray!40}-
            & \cellcolor{lightgray!40}-
            &\cellcolor{lightgray!40}\textcolor[RGB]{0,0,255}{\textbf{+5.60}}
            &\cellcolor{lightgray!40}\textcolor[RGB]{0,0,255}{\textbf{+2.99}}
            &\cellcolor{lightgray!40}\textcolor[RGB]{0,0,255}{\textbf{+2.14}}
            &\cellcolor{lightgray!40}\textcolor[RGB]{0,0,255}{\textbf{+4.94}}
            &\cellcolor{lightgray!40}\textcolor[RGB]{0,0,255}{\textbf{+2.07}}
            &\cellcolor{lightgray!40}\textcolor[RGB]{0,0,255}{\textbf{+1.00}}
            &\cellcolor{lightgray!40}\textcolor[RGB]{61,145,64}{{+10.60}}
            &\cellcolor{lightgray!40}\textcolor[RGB]{61,145,64} {{+12.89}}
            &\cellcolor{lightgray!40}\textcolor[RGB]{61,145,64} {{+16.62}} 
            &\cellcolor{lightgray!40}\textcolor[RGB]{0,0,255}{\textbf{+8.05}}
            &\cellcolor{lightgray!40}\textcolor[RGB]{0,0,255}{\textbf{+5.44}}
            &\cellcolor{lightgray!40}\textcolor[RGB]{0,0,255}{\textbf{+4.18}} \\ \hline
            
            \multirow{2}{*}{NetVLAD \cite{arandjelovic2016netvlad}} & \multirow{2}{*}{2016} & \multirow{2}{*}{V} & w/o & 88.26 & 92.49 & 93.73 & 94.09 & 97.41 & 98.43 & 2.01 & 4.58 & 6.88 & 83.68 & 94.04 & 96.23 \\
            
             &  &  & w & 90.85 & 94.29 & 95.49 & 95.44 & 98.19 & 98.96 & 15.76 & 22.06 & 29.80 & 89.64 & 96.34 & 97.80 \\
            
            \cellcolor{lightgray!40}$\Delta$
            & \cellcolor{lightgray!40}-
            & \cellcolor{lightgray!40}-
            & \cellcolor{lightgray!40}-
            &\cellcolor{lightgray!40}\textcolor[RGB]{61,145,64}{{+2.59}} 
            &\cellcolor{lightgray!40}\textcolor[RGB]{61,145,64}{{+1.80}} 
            &\cellcolor{lightgray!40}\textcolor[RGB]{61,145,64}{{+1.76}}
            &\cellcolor{lightgray!40}\textcolor[RGB]{61,145,64}{{+1.35}} 
            &\cellcolor{lightgray!40}\textcolor[RGB]{61,145,64}{{+0.78}} 
            &\cellcolor{lightgray!40}\textcolor[RGB]{61,145,64}{{+0.53}} 
            &\cellcolor{lightgray!40}\textcolor[RGB]{61,145,64}{{+13.75}}
            &\cellcolor{lightgray!40}\textcolor[RGB]{61,145,64} {{+17.48}}
            &\cellcolor{lightgray!40}\textcolor[RGB]{61,145,64} {{+22.92}} 
            &\cellcolor{lightgray!40}\textcolor[RGB]{61,145,64}{{+5.96}}
            &\cellcolor{lightgray!40}\textcolor[RGB]{61,145,64}{{+2.30}}
            &\cellcolor{lightgray!40}\textcolor[RGB]{61,145,64}{{+1.57}} \\ \hline
            
            \multirow{2}{*}{GeM \cite{radenovic2018fine}} & \multirow{2}{*}{2018} & \multirow{2}{*}{V} & w/o & 80.14 & 88.01 & 90.17 & 86.22 & 93.38 & 95.25 & 0.00 & 1.72 & 2.87 & 71.23 & 85.77 & 90.06 \\
            
             &  &  & w & 84.33 & 90.63 & 92.24 & 89.64 & 94.80 & 95.64  & 12.61 & 22.64 & 31.52 & 77.41 & 89.44 & 93.20 \\
             
            \cellcolor{lightgray!40}$\Delta$
            & \cellcolor{lightgray!40}-
            & \cellcolor{lightgray!40}-
            & \cellcolor{lightgray!40}-
            &\cellcolor{lightgray!40}\textcolor[RGB]{61,145,64}{{+4.19}} 
            &\cellcolor{lightgray!40}\textcolor[RGB]{61,145,64}{{+2.62}} 
            &\cellcolor{lightgray!40}\textcolor[RGB]{61,145,64}{{+2.07}}
            &\cellcolor{lightgray!40}\textcolor[RGB]{61,145,64}{{+3.42}} 
            &\cellcolor{lightgray!40}\textcolor[RGB]{61,145,64}{{+1.42}} 
            &\cellcolor{lightgray!40}\textcolor[RGB]{61,145,64}{{+0.39}} 
            &\cellcolor{lightgray!40}\textcolor[RGB]{61,145,64}{{+12.61}}
            &\cellcolor{lightgray!40}\textcolor[RGB]{61,145,64} {{+20.92}}
            &\cellcolor{lightgray!40}\textcolor[RGB]{61,145,64} {{+28.65}} 
            &\cellcolor{lightgray!40}\textcolor[RGB]{61,145,64}{{+6.18}}
            &\cellcolor{lightgray!40}\textcolor[RGB]{61,145,64}{{+3.67}}
            &\cellcolor{lightgray!40}\textcolor[RGB]{61,145,64}{{+3.14}} \\ \hline
            
            \multirow{2}{*}{Conv-AP \cite{ali2022gsv}} & \multirow{2}{*}{2022} & \multirow{2}{*}{V} & w/o & 88.93 & 93.87 & 94.80 & 94.48 & 98.62 & 99.16 & 4.58 & 8.88 & 12.03 & 84.83 & 94.98 & 97.49 \\
            
             &  &  & w & \textbf{91.40} & \textbf{95.53} & \textbf{96.85} & \textbf{95.67}  & \textbf{98.46} & 99.03 & \textbf{22.64} & \textbf{42.41} & \textbf{54.44} & 89.85 & \textbf{96.65} & \textbf{98.74} \\
             
            \cellcolor{lightgray!40}$\Delta$
            & \cellcolor{lightgray!40}-
            & \cellcolor{lightgray!40}-
            & \cellcolor{lightgray!40}-
            &\cellcolor{lightgray!40}\textcolor[RGB]{61,145,64}{{+2.47}} 
            &\cellcolor{lightgray!40}\textcolor[RGB]{61,145,64}{{+1.66}} 
            &\cellcolor{lightgray!40}\textcolor[RGB]{61,145,64}{{+2.05}}
            &\cellcolor{lightgray!40}\textcolor[RGB]{61,145,64}{{+1.19}} 
            &\cellcolor{lightgray!40}\textcolor[RGB]{61,145,64}{{-0.16}} 
            &\cellcolor{lightgray!40}\textcolor[RGB]{61,145,64}{{-0.13}} 
            &\cellcolor{lightgray!40}\textcolor[RGB]{0,0,255}{\textbf{+18.06}}
            &\cellcolor{lightgray!40}\textcolor[RGB]{0,0,255}{\textbf{+33.53}}
            &\cellcolor{lightgray!40}\textcolor[RGB]{0,0,255}{\textbf{+42.41}}
            &\cellcolor{lightgray!40}\textcolor[RGB]{61,145,64}{{+5.02}}
            &\cellcolor{lightgray!40}\textcolor[RGB]{61,145,64}{{+1.67}}
            &\cellcolor{lightgray!40}\textcolor[RGB]{61,145,64}{{+1.25}} \\ \hline
            
            \multirow{2}{*}{EigenPlaces \cite{berton2023eigenplaces}} & \multirow{2}{*}{2023} & \multirow{2}{*}{V} & w/o & 88.05 & 93.48 & 94.54 & 93.53 & 98.36 & 99.09 & 2.29 & 6.02 & 8.60 & 84.94 & 94.67 & 97.18 \\
            
             &  &  & w & 90.85 & 94.36 & 95.93 & 95.37 & 98.11 & 99.09 & 14.61 & 25.21 & 35.82 & \textbf{90.48} & 96.23 & 98.12 \\
             
            \cellcolor{lightgray!40}$\Delta$
            & \cellcolor{lightgray!40}-
            & \cellcolor{lightgray!40}-
            & \cellcolor{lightgray!40}-
            &\cellcolor{lightgray!40}\textcolor[RGB]{61,145,64}{{+2.80}} 
            &\cellcolor{lightgray!40}\textcolor[RGB]{61,145,64}{{+0.88}} 
            &\cellcolor{lightgray!40}\textcolor[RGB]{61,145,64}{{+1.39}}
            &\cellcolor{lightgray!40}\textcolor[RGB]{61,145,64}{{+1.84}} 
            &\cellcolor{lightgray!40}\textcolor[RGB]{61,145,64}{{-0.25}} 
            &\cellcolor{lightgray!40}\textcolor[RGB]{61,145,64}{{+0.00}} 
            &\cellcolor{lightgray!40}\textcolor[RGB]{61,145,64}{{+12.32}}
            &\cellcolor{lightgray!40}\textcolor[RGB]{61,145,64} {{+19.19}}
            &\cellcolor{lightgray!40}\textcolor[RGB]{61,145,64} {{+27.22}} 
            &\cellcolor{lightgray!40}\textcolor[RGB]{61,145,64}{{+5.54}}
            &\cellcolor{lightgray!40}\textcolor[RGB]{61,145,64}{{+1.56}}
            &\cellcolor{lightgray!40}\textcolor[RGB]{61,145,64}{{+0.94}} \\ \hline
            
            \multirow{2}{*}{MixVPR \cite{ali2023mixvpr}} & \multirow{2}{*}{2023} & \multirow{2}{*}{V} & w/o & 89.25 & 93.76 & 95.03 & 94.55 & 98.31 & 99.18 & 6.88 & 12.03 & 15.47 & 85.98 & 94.98 & 97.91 \\
            
             &  &  & w & 90.85 & 94.39 & 95.72 & 95.46 & 98.44 & \textbf{99.20} & 16.91 & 25.21 & 31.52 & 89.12 & 94.35 & 97.49 \\
             
            \cellcolor{lightgray!40}$\Delta$
            & \cellcolor{lightgray!40}-
            & \cellcolor{lightgray!40}-
            & \cellcolor{lightgray!40}-
            &\cellcolor{lightgray!40}\textcolor[RGB]{61,145,64}{{+1.60}} 
            &\cellcolor{lightgray!40}\textcolor[RGB]{61,145,64}{{+0.63}} 
            &\cellcolor{lightgray!40}\textcolor[RGB]{61,145,64}{{+0.69}}
            &\cellcolor{lightgray!40}\textcolor[RGB]{61,145,64}{{+0.91}} 
            &\cellcolor{lightgray!40}\textcolor[RGB]{61,145,64}{{+0.13}} 
            &\cellcolor{lightgray!40}\textcolor[RGB]{61,145,64}{{+0.02}} 
            &\cellcolor{lightgray!40}\textcolor[RGB]{61,145,64}{{+10.03}}
            &\cellcolor{lightgray!40}\textcolor[RGB]{61,145,64} {{+13.18}}
            &\cellcolor{lightgray!40}\textcolor[RGB]{61,145,64} {{+16.05}} 
            &\cellcolor{lightgray!40}\textcolor[RGB]{61,145,64}{{+3.14}}
            &\cellcolor{lightgray!40}\textcolor[RGB]{61,145,64}{{-0.63}}
            &\cellcolor{lightgray!40}\textcolor[RGB]{61,145,64}{{-0.42}} \\ \hline
        \end{tabular}
        \begin{tablenotes} 
            \item '$\Delta$' represents 'improvement'. Green numbers show our method's performance improvement over the baseline, while blue numbers indicate the maximum performance improvement among various methods. Bold black numbers highlight the best performance of different methods. These notations are consistent in \Cref{Tab.4} and \Cref{Tab.5}.
        \end{tablenotes} 
    \end{threeparttable}
    }
    \label{Tab.3} 
\vspace{-0.2cm}
\end{table*}

\begin{table*}[t]
\centering
\setlength{\abovecaptionskip}{-1pt} 
\caption{Ablation of Shared Bottom Backbone.}
\scalebox{0.89}{
\setlength{\tabcolsep}{6.5pt} 
\renewcommand{\arraystretch}{1.20} 
\begin{tabular}{c|c|c|ccc|ccc|ccc|ccc}
    \hline
    \# & \begin{tabular}[c]{@{}c@{}}Structural \\ Stream\end{tabular} & \begin{tabular}[c]{@{}c@{}}Bottom \\ Backbone \\ Pre-trained\end{tabular} & R@1   & R@5   & R@10  & R$^{\text{E}}$@1  & R$^{\text{E}}$@5  & R$^{\text{E}}$@10 & R$^{\text{H}}$@1  & R$^{\text{H}}$@5  & R$^{\text{H}}$@10 & R$^{\text{SH}}$@1 & R$^{\text{SH}}$@5 & R$^{\text{SH}}$@10 \\ \hline

    \ding{172} & \ding{55} & \ding{55} & 88.93 & 93.87 & 94.80 & 94.48 & 98.62 & 99.16 & 4.58  & 8.88  & 12.03 & 84.83 & 94.98 & 97.49  \\ \hline

    \ding{173} & \ding{55} & \checkmark & 88.75 & 93.73 & 94.90 & 93.70 & 97.79 & 98.50 & 10.89 & 19.48 & 26.93 & 86.51 & 95.61 & 97.28  \\
    \cellcolor{lightgray!40}$\Delta$
    &\cellcolor{lightgray!40}-
    & \cellcolor{lightgray!40}-
    &\cellcolor{lightgray!40}\textcolor[RGB]{61,145,64}{{-0.18}} 
    &\cellcolor{lightgray!40}\textcolor[RGB]{61,145,64}{{-0.14}} 
    &\cellcolor{lightgray!40}\textcolor[RGB]{61,145,64}{{+0.10}}
    &\cellcolor{lightgray!40}\textcolor[RGB]{61,145,64}{{-0.78}} 
    &\cellcolor{lightgray!40}\textcolor[RGB]{61,145,64}{{-0.83}} 
    &\cellcolor{lightgray!40}\textcolor[RGB]{61,145,64}{{-0.66}}
    &\cellcolor{lightgray!40}\textcolor[RGB]{61,145,64} {{+6.31}}
    &\cellcolor{lightgray!40}\textcolor[RGB]{61,145,64} {{+10.60}} 
    &\cellcolor{lightgray!40}\textcolor[RGB]{61,145,64}{{+14.90}}
    &\cellcolor{lightgray!40}\textcolor[RGB]{61,145,64}{{+1.68}}
    &\cellcolor{lightgray!40}\textcolor[RGB]{61,145,64}{{+0.63}}
    &\cellcolor{lightgray!40}\textcolor[RGB]{61,145,64}{{-0.21}}\\ \hline

    \ding{174} & \checkmark & \ding{55} & 90.59 & 94.03 & 95.45 & 95.24 & 98.21 & 99.14 & 15.19  & 22.64 & 29.23 & 89.12 & 94.04 & 96.55  \\
    \cellcolor{lightgray!40}$\Delta$
    &\cellcolor{lightgray!40}-
    & \cellcolor{lightgray!40}-
    &\cellcolor{lightgray!40}\textcolor[RGB]{61,145,64}{{+1.66}} 
    &\cellcolor{lightgray!40}\textcolor[RGB]{61,145,64}{{+0.16}} 
    &\cellcolor{lightgray!40}\textcolor[RGB]{61,145,64}{{+0.65}}
    &\cellcolor{lightgray!40}\textcolor[RGB]{61,145,64}{{+0.76}} 
    &\cellcolor{lightgray!40}\textcolor[RGB]{61,145,64}{{-0.41}} 
    &\cellcolor{lightgray!40}\textcolor[RGB]{61,145,64}{{-0.02}}
    &\cellcolor{lightgray!40}\textcolor[RGB]{61,145,64} {{+10.61}}
    &\cellcolor{lightgray!40}\textcolor[RGB]{61,145,64} {{+13.76}} 
    &\cellcolor{lightgray!40}\textcolor[RGB]{61,145,64}{{+17.20}}
    &\cellcolor{lightgray!40}\textcolor[RGB]{61,145,64}{{+4.29}}
    &\cellcolor{lightgray!40}\textcolor[RGB]{61,145,64}{{-0.94}}
    &\cellcolor{lightgray!40}\textcolor[RGB]{61,145,64}{{-0.94}}\\ \hline

    \ding{175} & \checkmark & \checkmark & 91.40 & 95.53 & 96.85 & 95.67 & 98.46 & 99.03 & 22.64 & 42.41 & 54.44 & 89.85 & 96.65 & 98.74  \\
    \cellcolor{lightgray!40}$\Delta$
    & \cellcolor{lightgray!40}-
    & \cellcolor{lightgray!40}-
    &\cellcolor{lightgray!40}\textcolor[RGB]{61,145,64}{{+2.47}} 
    &\cellcolor{lightgray!40}\textcolor[RGB]{61,145,64}{{+1.66}} 
    &\cellcolor{lightgray!40}\textcolor[RGB]{61,145,64}{{+2.05}}
    &\cellcolor{lightgray!40}\textcolor[RGB]{61,145,64}{{+1.19}} 
    &\cellcolor{lightgray!40}\textcolor[RGB]{61,145,64}{{-0.16}} 
    &\cellcolor{lightgray!40}\textcolor[RGB]{61,145,64}{{-0.13}}
    &\cellcolor{lightgray!40}\textcolor[RGB]{61,145,64} {{+18.06}}
    &\cellcolor{lightgray!40}\textcolor[RGB]{61,145,64} {{+33.53}} 
    &\cellcolor{lightgray!40}\textcolor[RGB]{61,145,64}{{+42.41}}
    &\cellcolor{lightgray!40}\textcolor[RGB]{61,145,64}{{+5.02}}
    &\cellcolor{lightgray!40}\textcolor[RGB]{61,145,64}{{+1.67}}
    &\cellcolor{lightgray!40}\textcolor[RGB]{61,145,64}{{+1.25}}\\ \hline
    \end{tabular}
}
\label{Tab.4} 
\vspace{-0.2cm}
\end{table*}

\begin{table*}[t]
\centering
\setlength{\abovecaptionskip}{-1pt} 
\caption{Ablation of BEV type.}
\scalebox{0.89}{
\setlength{\tabcolsep}{6.5pt} 
\renewcommand{\arraystretch}{1.20} 
\begin{tabular}{c|ccc|ccc|ccc|ccc|ccc}
\hline
\# & Div. & P.C. & Bound. & R@1   & R@5   & R@10  & R$^{\text{E}}$@1  & R$^{\text{E}}$@5  & R$^{\text{E}}$@10 & R$^{\text{H}}$@1  & R$^{\text{H}}$@5  & R$^{\text{H}}$@10 & R$^{\text{SH}}$@1 & R$^{\text{SH}}$@5 & R$^{\text{SH}}$@10 \\ \hline
\ding{172}  & -       & -            & -        & 88.93 & 93.87 & 94.80 & 94.48 & 98.62 & 99.16 & 4.58  & 8.88  & 12.03 & 84.83 & 94.98  & 97.49   \\
\ding{173}  & \checkmark      & -            & -        & 91.25 & 94.90 & 96.05 & \textbf{96.09} & 98.83 & 99.41 & 14.33 & 25.50 & 32.95 & 89.12 & 95.71  & 98.12   \\
\ding{174}  & -       & \checkmark           & -        & 89.96 & 94.21 & 95.56 & 95.19 & 98.36 & 99.08 & 9.46  & 22.35 & 30.66 & 86.72 & 94.56  & 97.28   \\
\ding{175}  & -       & -            & \checkmark       & 90.11 & 94.90 & 96.20 & 94.94 & \textbf{98.84} & 99.46 & 15.76 & 24.93 & 34.96 & 87.13 & 95.82  & 98.22   \\
\ding{176}  & \checkmark      & \checkmark           & -        & 91.02 & 95.05 & 96.30 & \textbf{96.09} & 98.59 & 99.30 & 14.33 & 30.09 & 39.54 & 87.34 & 96.65  & 98.33   \\
\ding{177}  & \checkmark      & -            & \checkmark       & 91.11 & 94.65 & 95.82 & 95.92 & 98.59 & 99.20 & 17.48 & 26.07 & 33.24 & 87.97 & 95.08  & 97.59   \\
\ding{178}  & -       & \checkmark           & \checkmark       & 90.77 & 94.69 & 95.68 & 95.79 & \textbf{98.84} & \textbf{99.48} & 9.74  & 17.19 & 23.21 & 89.02 & \textbf{97.07}  & 98.43   \\
\ding{179}  & \checkmark      & \checkmark           & \checkmark       & \textbf{91.40} & \textbf{95.53} & \textbf{96.85} & 95.67 & 98.46 & 99.03 & \textbf{22.64} & \textbf{42.41} & \textbf{54.44} & \textbf{89.85} & 96.65  & \textbf{98.74}   \\ \hline
\end{tabular}
}
\label{Tab.5} 
\vspace{-16pt}
\end{table*}

\textbf{Evaluation.} The top-$k$ ($k = 1, 5, 10$) recall is adopted as the evaluation metric, based on which, we evaluate our framework on the whole dataset as well as three subsets with different levels of recall difficulty, \ie, R@$k$ (whole), R$^{\text{E}}$@$k$ (easy), R$^{\text{H}}$@$k$ (hard), R$^{\text{SH}}$@$k$ (semi-hard). To specify, we first calculate the similarity $d_{ij}$ between sample $s_i$ and sample $s_j$ ($j \neq i$) in the test dataset/subset, based on which, the index of $n$ most similar samples is obtained, among which, we then exclude consecutive frames and detect $k$ $(k \ll n)$ positive samples, indicating successful recall.

\begin{figure}[t] 
\centering 
\setlength{\abovecaptionskip}{-2pt} 
\includegraphics[width=0.48\textwidth]{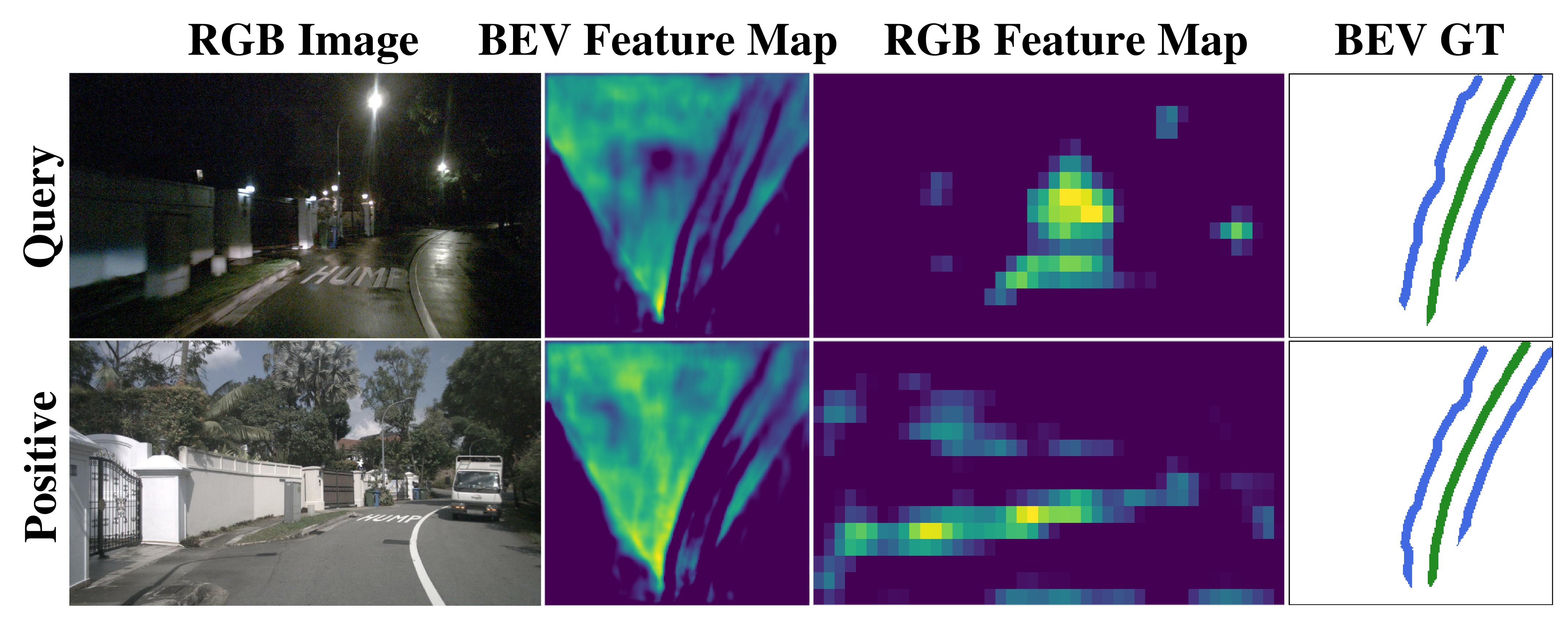} 
\caption{Visualization comparisons of RGB and BEV features. Obviously, BEV features exhibit more structural characteristics than RGB features.} 
\label{Fig.4} 
\vspace{-0.6cm}
\end{figure}

\subsection{Evaluation for Visual Place Recognition}
We apply six baseline models to our framework and evaluate the VPR performance based on monocular images. \Cref{Tab.3} shows the recall results on VPR-NuScenes. All methods incorporating structural streams have shown improvements in almost all metrics, with the absolute gains on R@1 over the whole dataset being 5.60\%, 2.59\%, 4.19\%, 2.47\%, 2.80\%, 1.60\%. From the results on other subsets, it can also be observed that datasets do not limit BEV$^2$PR: on datasets with remarkable appearance differences, it can effectively improve performance (refer to next two paragraphs); on datasets with insignificant appearance differences, BEV$^2$PR maintains original RGB performance. Note that we train all models incorporating structural streams with the same hyper-parameters (\eg, loss function coefficients). This may have prevented some models from achieving optimal performance, potentially explaining why MixVPR with the structural stream under-performs compared to ConvAP. However, this does not detract the overall performance improvements.

\textbf{Illumination Change.} Compared to the easy set, the performance improvement on the hard set is more significant, with a maximum increase of 18.06\% on R$^{\text{H}}$@1. This can be attributed to the fact that in scenes with significant illumination changes, the semantic and texture information becomes completely ineffective and structural features in the BEV frame are relatively stable. As shown in the second row of \Cref{Fig.3}, it is difficult for human eyes to distinguish the complete scene, while just like in the nighttime scenes of autonomous driving in the real world, grounds are particularly clear compared to other areas, especially lane boundary and lane divider. \Cref{Fig.4} presents the feature maps of RGB and BEV during the day and night, respectively. It is evident that under such conditions, the effectiveness of structural information in the BEV frame is far greater than that of appearance information in the RGB frame.

\textbf{Image Clarity Change.} The methods with added structural cues can also yield considerable performance enhancements in scenes where there is a substantial change in image clarity, with an average improvement of 5.65\% on R$^{\text{SH}}$@1 across different models. As shown in the third row of \Cref{Fig.3}, compared with positive images, the query image is more blurry, but the overall image content remains basically unchanged. The semantic and texture information in certain areas is relatively complete and the structural information can further assist in generating more comprehensive features.

\subsection{Ablation Study}


\textbf{Shared Bottom Backbone.} To demonstrate the effectiveness of the shared bottom backbone, we intentionally degenerate our framework into two versions, as depicted in \Cref{Tab.4}, one (\ding{173}) is single-stream with a bottom backbone pre-trained through BEV generation, the other (\ding{174}) is dual-stream with a bottom backbone not pre-trained via BEV generation. The results reveal that the pre-trained bottom backbone helps visual stream learn fine-grained local features (\ding{172} \vs \ding{173}), especially map information such as lane divider, pedestrian crossing, and lane boundary in our work. This is more effective for hard scenes, but to some extent, it affects the effectiveness of global features in simple scenes. For \ding{174}, the dual-stream network can provide more detailed BEV structural information, thereby resulting in greater performance improvements in hard and semi-hard scenes. Based on the above discussion, the dual-stream network with a shared bottom backbone can maximize the robustness of global features under different environments. 

\textbf{BEV Types.} We then perform ablation experiments to study the impact of varying types of BEV supervised models on VPR. Given the restricted variety of map elements within the NuScenes dataset, we focus on three key static elements: lane divider, pedestrian crossing, and lane boundary. As depicted in \Cref{Tab.5}, when training without structural information, the baseline model achieves 88.93\% R@1 and 4.58\% R$^{\text{H}}$@1, respectively. When providing structural information of lane divider to the model, BEV$^2$PR achieves gains of 2.32\% and 9.75\% on R@1 and R$^{\text{H}}$@1. More types of BEV consistently improves our method, particularly benefiting the categories of lane divider and lane boundary, which can be attributed to the higher proportion of road divider and road boundary in the image compared to pedestrian crossing. While three types of static elements with structural characteristics can achieve the best performance improvement.

\begin{table}[]
\centering
\vspace{6pt}
\setlength{\abovecaptionskip}{-1pt} 
\caption{Generalization Performance.}
\setlength{\tabcolsep}{8pt} 
\scalebox{0.95}{
    \begin{threeparttable}
        \begin{tabular}{lccccc}
            \hline
            Sequence           & 00   & 02   & 05   & 06   & Mean \\ \hline
            Vision             & 93.7 & 75.2 & 84.3 & 95.6 & 87.2 \\
            Vision + Structure & 94.4 & 78.1 & 85.9 & 98.2 & 89.2 \\ \hline
        \end{tabular}
    \end{threeparttable}
}
\label{Tab.6}
\vspace{-12pt}
\end{table}

\subsection{Generalization.} We also evaluate the VPR performance of our methods on the KITTI dataset \cite{geiger2012we} utilizing models trained on VPR-NuScenes. The dataset partition of KITTI is the same as \cite{luo2023bevplace}. We implement BEV$^2$PR with MixVPR as the aggregation module of the visual stream. \Cref{Tab.6} presents the R$^{\text{E}}$@1 on the validation set. The method with added structural cues demonstrates more robust performance, although the BEV features may not be completely accurate.

\section{CONCLUSIONS}

In this paper, we propose a new BEV-enhanced VPR framework by exploiting the structural cues in BEV from a single monocular camera and introduce a new dataset for place recognition. The cores of our success lie in the introduction of BEV structural information and the promotion of the shared bottom backbone for local feature learning in the visual stream. Analytical experiments exhibit the superiority of BEV structural information in retrieving hard samples. We hope our work can enlighten more research on VPR. In future work, we will explore the impact of additional BEV categories on VPR performance and lightweight our dual-stream model through distillation techniques.

\section*{Acknowledgments}

This work was supported in part by the Beijing Natural Science Foundation (Grant No. L223003, JQ22014), the Natural Science Foundation of China (Grant No. U22B2056, 62192782, 62036011, U2033210, 62102417), the Project of Beijing Science and technology Committee (Project No. Z231100005923046). Jin Gao was also supported in part by the Youth Innovation Promotion Association, CAS.


\bibliographystyle{IEEEtran}
\bibliography{reference}

\begin{thebibliography}{10}
\providecommand{\url}[1]{#1}
\csname url@samestyle\endcsname
\providecommand{\newblock}{\relax}
\providecommand{\bibinfo}[2]{#2}
\providecommand{\BIBentrySTDinterwordspacing}{\spaceskip=0pt\relax}
\providecommand{\BIBentryALTinterwordstretchfactor}{4}
\providecommand{\BIBentryALTinterwordspacing}{\spaceskip=\fontdimen2\font plus
\BIBentryALTinterwordstretchfactor\fontdimen3\font minus \fontdimen4\font\relax}
\providecommand{\BIBforeignlanguage}[2]{{%
\expandafter\ifx\csname l@#1\endcsname\relax
\typeout{** WARNING: IEEEtran.bst: No hyphenation pattern has been}%
\typeout{** loaded for the language `#1'. Using the pattern for}%
\typeout{** the default language instead.}%
\else
\language=\csname l@#1\endcsname
\fi
#2}}
\providecommand{\BIBdecl}{\relax}
\BIBdecl

\bibitem{lowry2015visual}
S.~Lowry, N.~S{\"u}nderhauf, P.~Newman, J.~J. Leonard, D.~Cox, P.~Corke, and M.~J. Milford, ``Visual place recognition: A survey,'' \emph{IEEE Transactions on Robotics}, vol.~32, no.~1, pp. 1--19, 2015.

\bibitem{masone2021survey}
C.~Masone and B.~Caputo, ``A survey on deep visual place recognition,'' \emph{IEEE Access}, vol.~9, pp. 19\,516--19\,547, 2021.

\bibitem{zhang2021visual}
X.~Zhang, L.~Wang, and Y.~Su, ``Visual place recognition: A survey from deep learning perspective,'' \emph{Pattern Recognition}, vol. 113, p. 107760, 2021.

\bibitem{luo2023bevplace}
L.~Luo, S.~Zheng, Y.~Li, Y.~Fan, B.~Yu, S.-Y. Cao, J.~Li, and H.-L. Shen, ``Bevplace: Learning lidar-based place recognition using bird's eye view images,'' in \emph{Proceedings of the IEEE/CVF International Conference on Computer Vision}, 2023, pp. 8700--8709.

\bibitem{komorowski2021minkloc++}
J.~Komorowski, M.~Wysocza{\'n}ska, and T.~Trzcinski, ``Minkloc++: lidar and monocular image fusion for place recognition,'' in \emph{International Joint Conference on Neural Networks}, 2021, pp. 1--8.

\bibitem{lai2022adafusion}
H.~Lai, P.~Yin, and S.~Scherer, ``Adafusion: Visual-lidar fusion with adaptive weights for place recognition,'' \emph{IEEE Robotics and Automation Letters}, vol.~7, no.~4, pp. 12\,038--12\,045, 2022.

\bibitem{zhou2023lcpr}
Z.~Zhou, J.~Xu, G.~Xiong, and J.~Ma, ``Lcpr: A multi-scale attention-based lidar-camera fusion network for place recognition,'' \emph{IEEE Robotics and Automation Letters}, 2023.

\bibitem{hu2020dasgil}
H.~Hu, Z.~Qiao, M.~Cheng, Z.~Liu, and H.~Wang, ``Dasgil: Domain adaptation for semantic and geometric-aware image-based localization,'' \emph{IEEE Transactions on Image Processing}, vol.~30, pp. 1342--1353, 2020.

\bibitem{oertel2020augmenting}
A.~Oertel, T.~Cieslewski, and D.~Scaramuzza, ``Augmenting visual place recognition with structural cues,'' \emph{IEEE Robotics and Automation Letters}, vol.~5, no.~4, pp. 5534--5541, 2020.

\bibitem{shen2023structvpr}
Y.~Shen, S.~Zhou, J.~Fu, R.~Wang, S.~Chen, and N.~Zheng, ``Structvpr: Distill structural knowledge with weighting samples for visual place recognition,'' in \emph{Proceedings of the IEEE/CVF Conference on Computer Vision and Pattern Recognition}, 2023, pp. 11\,217--11\,226.

\bibitem{kirillov2023segment}
A.~Kirillov, E.~Mintun, N.~Ravi, H.~Mao, C.~Rolland, L.~Gustafson, T.~Xiao, S.~Whitehead, A.~C. Berg, W.-Y. Lo \emph{et~al.}, ``Segment anything,'' \emph{arXiv preprint arXiv:2304.02643}, 2023.

\bibitem{arandjelovic2016netvlad}
R.~Arandjelovic, P.~Gronat, A.~Torii, T.~Pajdla, and J.~Sivic, ``Netvlad: Cnn architecture for weakly supervised place recognition,'' in \emph{Proceedings of the IEEE/CVF Conference on Computer Vision and Pattern Recognition}, 2016, pp. 5297--5307.

\bibitem{yu2019spatial}
J.~Yu, C.~Zhu, J.~Zhang, Q.~Huang, and D.~Tao, ``Spatial pyramid-enhanced netvlad with weighted triplet loss for place recognition,'' \emph{IEEE Transactions on Neural Networks and Learning Systems}, vol.~31, no.~2, pp. 661--674, 2019.

\bibitem{zhang2021vector}
J.~Zhang, Y.~Cao, and Q.~Wu, ``Vector of locally and adaptively aggregated descriptors for image feature representation,'' \emph{Pattern Recognition}, vol. 116, p. 107952, 2021.

\bibitem{radenovic2018fine}
F.~Radenovi{\'c}, G.~Tolias, and O.~Chum, ``Fine-tuning cnn image retrieval with no human annotation,'' \emph{IEEE Transactions on Pattern Analysis and Machine Intelligence}, vol.~41, no.~7, pp. 1655--1668, 2018.

\bibitem{berton2022rethinking}
G.~Berton, C.~Masone, and B.~Caputo, ``Rethinking visual geo-localization for large-scale applications,'' in \emph{Proceedings of the IEEE/CVF Conference on Computer Vision and Pattern Recognition}, 2022, pp. 4878--4888.

\bibitem{wang2022transvpr}
R.~Wang, Y.~Shen, W.~Zuo, S.~Zhou, and N.~Zheng, ``Transvpr: Transformer-based place recognition with multi-level attention aggregation,'' in \emph{Proceedings of the IEEE/CVF Conference on Computer Vision and Pattern Recognition}, 2022, pp. 13\,648--13\,657.

\bibitem{ali2022gsv}
A.~Ali-bey, B.~Chaib-draa, and P.~Gigu{\`e}re, ``Gsv-cities: Toward appropriate supervised visual place recognition,'' \emph{Neurocomputing}, vol. 513, pp. 194--203, 2022.

\bibitem{lu2020pic}
Y.~Lu, F.~Yang, F.~Chen, and D.~Xie, ``Pic-net: Point cloud and image collaboration network for large-scale place recognition,'' \emph{arXiv preprint arXiv:2008.00658}, 2020.

\bibitem{ratz2020oneshot}
S.~Ratz, M.~Dymczyk, R.~Siegwart, and R.~Dub{\'e}, ``Oneshot global localization: Instant lidar-visual pose estimation,'' in \emph{IEEE International conference on Robotics and Automation}, 2020, pp. 5415--5421.

\bibitem{pan2021coral}
Y.~Pan, X.~Xu, W.~Li, Y.~Cui, Y.~Wang, and R.~Xiong, ``Coral: Colored structural representation for bi-modal place recognition,'' in \emph{IEEE/RSJ International Conference on Intelligent Robots and Systems}, 2021, pp. 2084--2091.

\bibitem{qin2021structure}
C.~Qin, Y.~Zhang, Y.~Liu, D.~Zhu, S.~A. Coleman, and D.~Kerr, ``Structure-aware feature disentanglement with knowledge transfer for appearance-changing place recognition,'' \emph{IEEE Transactions on Neural Networks and Learning Systems}, 2021.

\bibitem{philion2020lift}
J.~Philion and S.~Fidler, ``Lift, splat, shoot: Encoding images from arbitrary camera rigs by implicitly unprojecting to 3d,'' in \emph{European Conference on Computer Vision}, 2020, pp. 194--210.

\bibitem{wu2021mobilesal}
Y.~Wu, Y.~Liu, J.~Xu, J.~Bian, Y.~Gu, and M.~Cheng, ``Mobilesal: Extremely efficient rgb-d salient object detection,'' \emph{IEEE Transactions on Pattern Analysis and Machine Intelligence}, vol.~44, no.~12, pp. 10\,261--10\,269, 2021.

\bibitem{7410507}
A.~B. Yandex and V.~Lempitsky, ``Aggregating local deep features for image retrieval,'' in \emph{IEEE International Conference on Computer Vision}, 2015, pp. 1269--1277.

\bibitem{berton2023eigenplaces}
G.~Berton, G.~Trivigno, B.~Caputo, and C.~Masone, ``Eigenplaces: Training viewpoint robust models for visual place recognition,'' in \emph{Proceedings of the IEEE/CVF International Conference on Computer Vision}, 2023, pp. 11\,080--11\,090.

\bibitem{ali2023mixvpr}
A.~Ali-Bey, B.~Chaib-Draa, and P.~Giguere, ``Mixvpr: Feature mixing for visual place recognition,'' in \emph{Proceedings of the IEEE/CVF Winter Conference on Applications of Computer Vision}, 2023, pp. 2998--3007.

\bibitem{li2022hdmapnet}
Q.~Li, Y.~Wang, Y.~Wang, and H.~Zhao, ``Hdmapnet: An online hd map construction and evaluation framework,'' in \emph{International Conference on Robotics and Automation}, 2022, pp. 4628--4634.

\bibitem{caesar2020nuscenes}
H.~Caesar, V.~Bankiti, A.~H. Lang, S.~Vora, V.~E. Liong, Q.~Xu, A.~Krishnan, Y.~Pan, G.~Baldan, and O.~Beijbom, ``nuscenes: A multimodal dataset for autonomous driving,'' in \emph{Proceedings of the IEEE/CVF Conference on Computer Vision and Pattern Recognition}, 2020, pp. 11\,621--11\,631.

\bibitem{ge2020self}
Y.~Ge, H.~Wang, F.~Zhu, R.~Zhao, and H.~Li, ``Self-supervising fine-grained region similarities for large-scale image localization,'' in \emph{European Conference on Computer Vision}, 2020, pp. 369--386.

\bibitem{geiger2012we}
A.~Geiger, P.~Lenz, and R.~Urtasun, ``Are we ready for autonomous driving? the kitti vision benchmark suite,'' in \emph{Proceedings of the IEEE/CVF Conference on Computer Vision and Pattern Recognition}, 2012, pp. 3354--3361.

\end{thebibliography}

\addtolength{\textheight}{-12cm}   

\end{document}